\documentclass[runningheads]{llncs}

\usepackage[mobile]{eccv}

\usepackage{eccvabbrv}

\usepackage{graphicx}
\usepackage{booktabs}

\usepackage[accsupp]{axessibility}  

\usepackage{hyperref}

\usepackage{orcidlink}

\begin{document}

\title{Learning 3D Affordances for Blade Insertion in Cluttered Stowing} 

\titlerunning{Learning 3D Blade Insertion}

\author{Tianyu Li \and
Harpreet Sawhney \and
Minju Jung \and
Aditya Mehrotra \and
Kunal Mehrotra \and
Mudit Agrawal}

\authorrunning{T. Li et al.}

\institute{Amazon Robotics, USA \\
\email{Corresponding author: hasawhne@amazon.com}}

\maketitle



\newcommand{\figBladeInsertion}{%
\begin{figure}[hbt]
    \vspace{-14pt}
    \centering
    \includegraphics[width=0.8\linewidth]{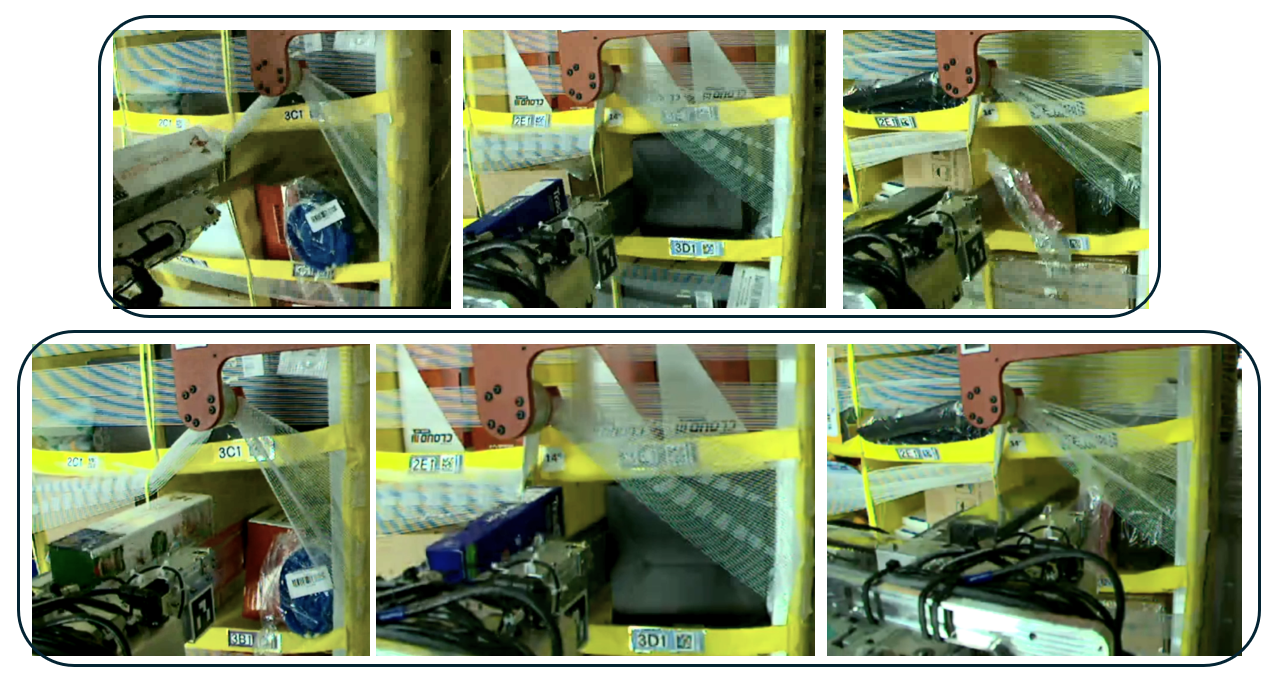}
    \caption{\textbf{Blade insertion for stowing.} \textbf{Top:} Initial bin state, blade approach, 
    and insertion into a densely cluttered bin. \textbf{Bottom:} 
    Blade navigating tight gaps while sweeping. Precise 3D geometric reasoning avoids damaging items/bin walls while reaching sufficiently deep to start sweeping items to create space.}
    \vspace{-18pt}  
    \label{fig:blade_insertion}
\end{figure}
}

\newcommand{\figBladePredResults}{%
\begin{figure}[hbt]
    \vspace{-14pt}
    \centering
    \includegraphics[width=\linewidth]{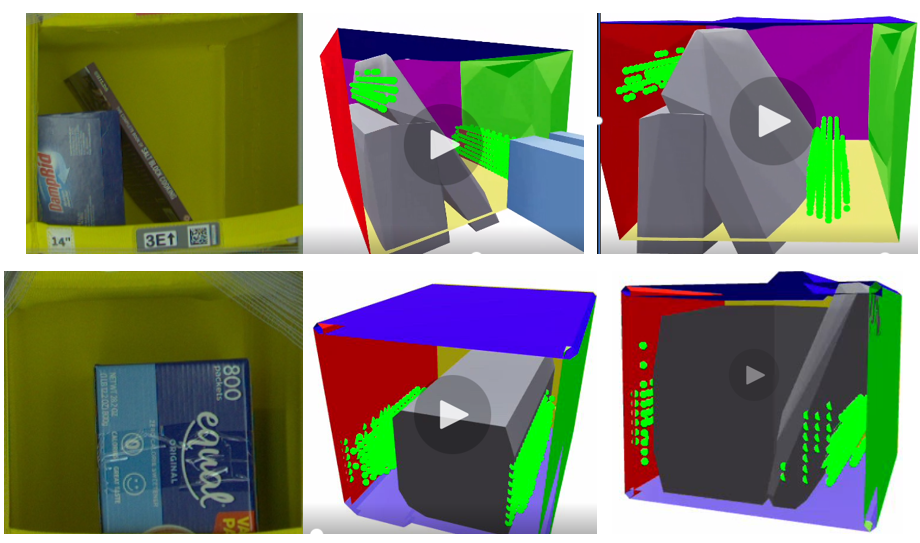}
    \caption{\textbf{VulcanVoxel Blade Prediction Visuals.} Shown as green dots representing blade voxel occupancy within a rendering of the reconstructed bin-item 3D mesh. Bin walls are colored, items are in gray. \textbf{Top:} Bin image, Blade occupancy predictions for this image and another sample bin. \textbf{Bottom:} 
    Vulcan Distill predictions for bin on the left and another sample bin.}
    \vspace{-18pt}  
    \label{fig:blade_pred_results}
\end{figure}
}

\newcommand{\figVulcanvoxelPipeline}{%
\begin{figure}[t]
    \centering
    \includegraphics[width=1.0\linewidth,trim={0cm 0cm 0cm 0},clip]{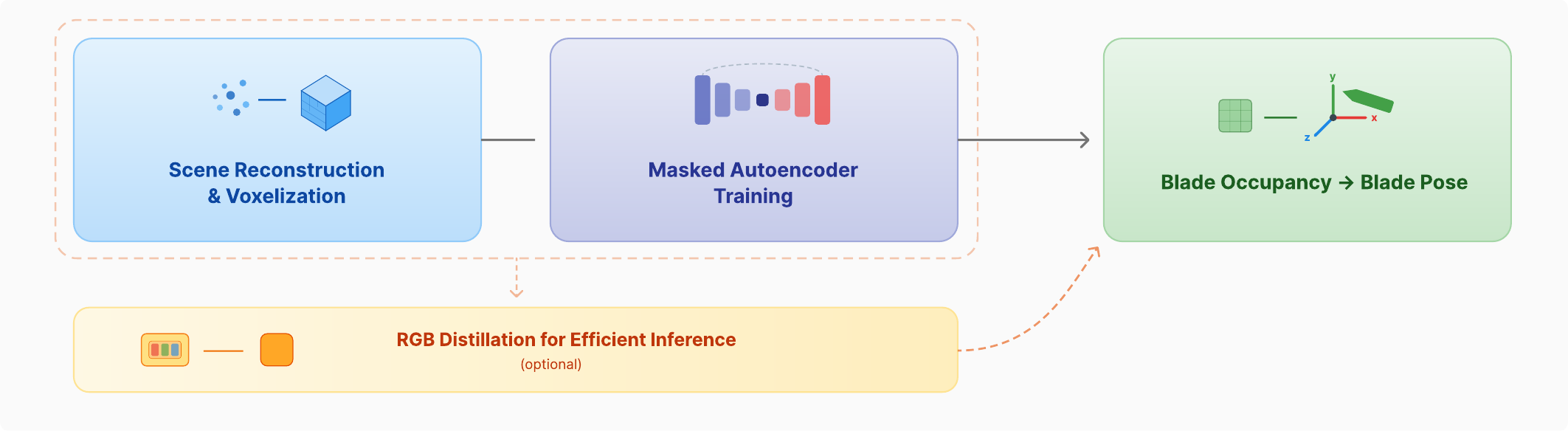}
    \caption{\textbf{VulcanVoxel pipeline.} Depth observations are voxelized into multi-channel occupancy grids, then a masked autoencoder reconstructs blade occupancy
  from scene context. Predicted occupancy is converted to blade pose via geometric scoring. An optional RGB distillation path (bottom) replaces the explicit stages for
  efficient inference.}
    \label{fig:vulcanvoxel_pipeline}
\end{figure}
}

\newcommand{\figVulcanvoxelArch}{%
\begin{figure}[t]
    \centering
    \includegraphics[width=1.0\linewidth,trim={0cm 0cm 0cm 0},clip]{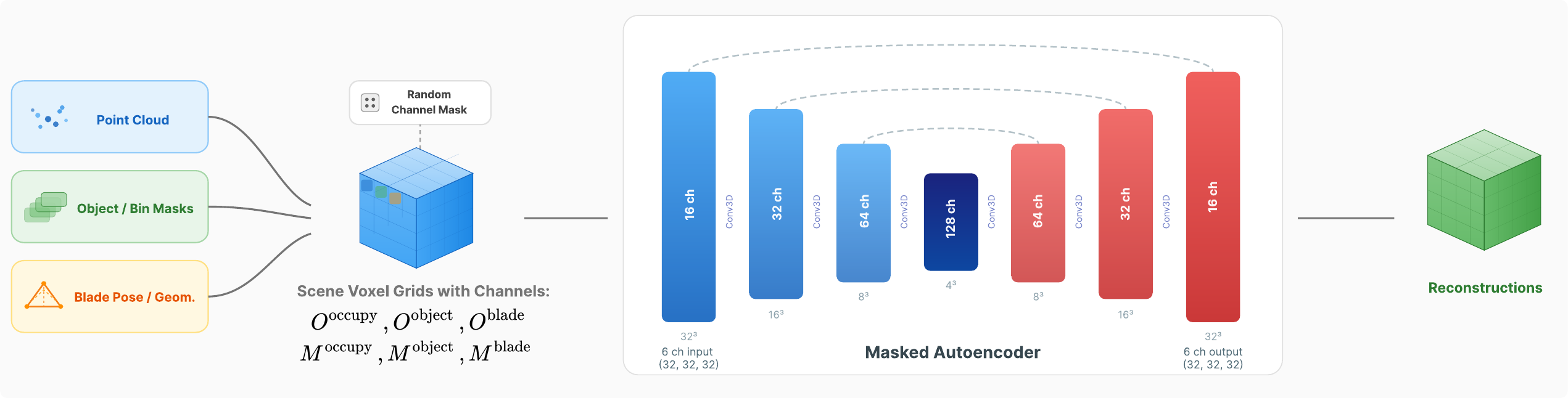}
    \caption{\textbf{VulcanVoxel architecture.} Point cloud, object/bin masks, and blade geometry are used to construct a multi-channel 3D occupancy grid. During
  training, random channel masking is applied before the grid is passed through a 3D UNet autoencoder that reconstructs the masked channels.}
    \label{fig:vulcanvoxel_arch}
\end{figure}
}

\newcommand{\figTabGenPlusTabGCU}{%
\begin{figure}[t]
    \centering
    \includegraphics[width=1.0\linewidth,trim={0cm 0cm 0cm 0},clip]{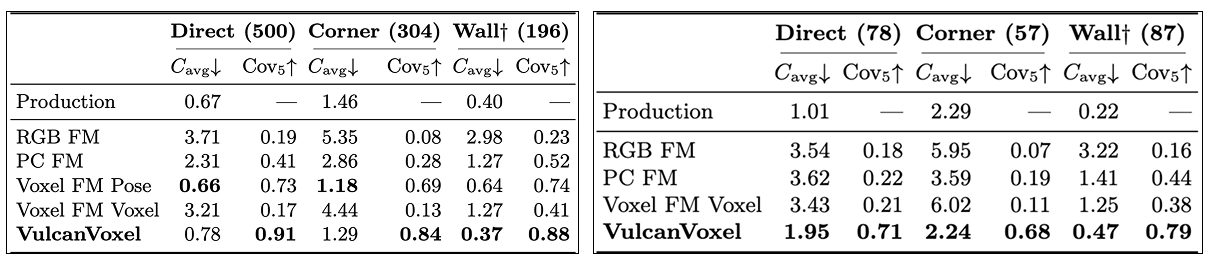}
    \caption{\textbf{Left:} Generalization to August 2025  (1,000 cycles). Method rankings preserved across the two-month temporal gap. \textbf{Right:} High clutter (GCU $>$ 50\%, 
July 2025). VulcanVoxel degrades less in 
coverage than competing methods; the 
$\text{Cov}_5$ advantage over Voxel FM Pose 
grows under clutter. ($\dagger$Wall Insert 
omits reach cost.)}
\label{fig:TabGenPlusTabGCU}
\end{figure}
}

\newcommand{\figChallDistill}{
\begin{figure}[t]
    \centering
    \includegraphics[width=1.0\linewidth,trim={0cm 0cm 0cm 0},clip]{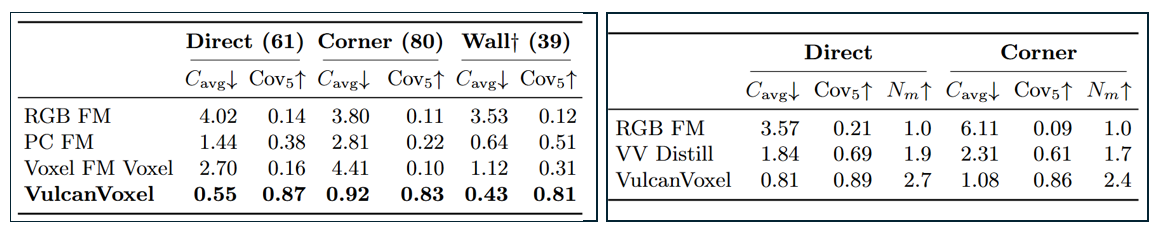}
    \caption{\textbf{Left:} Challenging configurations (production failures, August 2025). VulcanVoxel maintains coverage consistent with the successful test set while all baselines degrade substantially. Results represent a generalization test on difficult bin configurations, not a failure correction benchmark. \textbf{Right:} Distillation quality, July 2025. VV Distill substantially outperforms RGB FM at comparable speed, preserving partial multi-modality and $\sim 65\%$ of teacher
prediction quality.}
\label{fig:TabChallPlusTabDistill}
\end{figure}
}

\newcommand{\tabMainResults}{%
\begin{table}[t]
\centering
\caption{\textbf{Operational cost and top-5 
coverage, July 2025 (1,000 cycles).} 
$C_{\text{avg}}$: mean single-prediction cost. 
Viol: hard violations (any mesh penetration 
exceeding safety margin $m_e\!=\!0.01$m). 
$\text{Cov}_5$: fraction of scenes with at 
least one valid pose among top 5. 
Production (Prod) generates one prediction 
per scene; $\text{Cov}_5$ not applicable. 
$\dagger$Wall Insert omits reach cost.}
\label{tab:main_results}
\setlength{\tabcolsep}{2.5pt}
\small
\begin{tabular}{lrrrrrrrrrr}
\toprule
& \multicolumn{3}{c}{\textbf{Direct (508)}} 
& \multicolumn{3}{c}{\textbf{Corner (288)}} 
& \multicolumn{2}{c}{\textbf{Wall$\dagger$ (204)}} 
& \multicolumn{2}{c}{\textbf{All}} \\
\cmidrule(lr){2-4}\cmidrule(lr){5-7}
\cmidrule(lr){8-9}\cmidrule(lr){10-11}
& $C$$\downarrow$ & Viol$\downarrow$ 
& $\text{Cov}_5$$\uparrow$ 
& $C$$\downarrow$ & Viol$\downarrow$ 
& $\text{Cov}_5$$\uparrow$
& $C$$\downarrow$ & Viol$\downarrow$  
& $C$$\downarrow$ & Viol$\downarrow$ \\
\midrule
Production & 0.63 & 0 & --- 
& 1.39 & 0 & --- 
& 0.77 & 0 
& 0.88 & 0 \\
\midrule
RGB FM 
& 3.57 & 207 & 0.21
& 6.11 & 223 & 0.09
& 2.68 & 110 
& 4.27 & 540 \\
PC FM 
& 2.08 & 58 & 0.44
& 2.72 & 53 & 0.31
& 1.15 & 14 
& 2.06 & 125 \\
Voxel FM Pose 
& \textbf{0.60} & \textbf{3} & 0.71
& 1.18 & \textbf{0} & 0.68
& 0.25 & \textbf{0} 
& \textbf{0.72} & \textbf{3} \\
Voxel FM Voxel 
& 3.13 & 100 & 0.19
& 4.45 & 75 & 0.14
& 1.23 & 20 
& 2.99 & 195 \\
\textbf{VulcanVoxel} 
& 0.81 & 5 & \textbf{0.89}
& \textbf{1.08} & 4 & \textbf{0.86}
& \textbf{0.42} & 2
& 0.79 & 11 \\
\bottomrule
\end{tabular}
\end{table}
}

\newcommand{\tabAugResults}{%
\begin{table}[t]
\centering
\caption{\textbf{Generalization to August 2025 
(1,000 cycles).} Method rankings preserved 
across the two-month temporal gap. 
$\dagger$Wall Insert omits reach cost.}
\label{tab:aug_results}
\setlength{\tabcolsep}{2.5pt}
\small
\begin{tabular}{lrrrrrr}
\toprule
& \multicolumn{2}{c}{\textbf{Direct (500)}} 
& \multicolumn{2}{c}{\textbf{Corner (304)}} 
& \multicolumn{2}{c}{\textbf{Wall$\dagger$ (196)}}\\
\cmidrule(lr){2-3}\cmidrule(lr){4-5}
\cmidrule(lr){6-7}
& $C_{\text{avg}}$$\downarrow$ 
& $\text{Cov}_5$$\uparrow$
& $C_{\text{avg}}$$\downarrow$ 
& $\text{Cov}_5$$\uparrow$
& $C_{\text{avg}}$$\downarrow$ 
& $\text{Cov}_5$$\uparrow$ \\
\midrule
Production & 0.67 & --- & 1.46 & --- & 0.40 & --- \\
\midrule
RGB FM & 3.71 & 0.19 & 5.35 & 0.08 
& 2.98 & 0.23 \\
PC FM & 2.31 & 0.41 & 2.86 & 0.28 
& 1.27 & 0.52 \\
Voxel FM Pose & \textbf{0.66} & 0.73 & \textbf{1.18} & 0.69 
& 0.64 & 0.74 \\
Voxel FM Voxel & 3.21 & 0.17 & 4.44 & 0.13 
& 1.27 & 0.41 \\
\textbf{VulcanVoxel} & 0.78
& \textbf{0.91} & 1.29
& \textbf{0.84} & \textbf{0.37} 
& \textbf{0.88} \\
\bottomrule
\end{tabular}
\end{table}
}

\newcommand{\tabStrategies}{%
\begin{table}[h]
\vspace{-4pt}
\centering
\caption{\textbf{Insertion strategies} with 
critical perceptual pose as the affordance 
prediction target. Motion primitives: 
\textsc{Approach} (move to bin opening), 
\textsc{ExtendBlade} (extend blade inside bin), 
\textsc{ItemPush} (push the item inside to create initial clearance to seek the wall next to the item.), 
\textsc{WallSeek} (seek wall via force), 
\textsc{FollowWall} (translate along wall), 
\textsc{LipSeek} (seek bin lip, a stiff band at the bin bottom).}
\label{tab:strategies}
\setlength{\tabcolsep}{4pt}
\small
\begin{tabular}{lll}
\toprule
\textbf{Strategy} & 
\textbf{Action Sequence} & 
\textbf{Critical Pose} \\
\midrule
Direct & 
\textsc{Approach} $\to$ 
\textsc{ExtendBlade} & 
\textsc{ExtendBlade} \\
Corner & 
\textsc{Approach} $\to$ 
\textsc{ExtendBlade} $\to$ & 
\textsc{ExtendBlade} \\
& \textsc{WallSeek} $\to$ 
\textsc{(Bin)LipSeek} & \\
Wall & 
\textsc{Approach} $\to$ 
(\textsc{ItemPush}) $\to$ & 
\textsc{WallSeek} \\
& \textsc{WallSeek} $\to$ 
\textsc{FollowWall} $\to$
\textsc{ExtendBlade} & \\
\bottomrule
\end{tabular}
\vspace{-4pt}
\end{table}
}

\newcommand{\tabModels}{%
\begin{table}[t]
\centering
\caption{\textbf{Methods.} VV Distill is evaluated 
separately in Sec.~\ref{sec:experiments_distill}.}
\label{tab:models}
\setlength{\tabcolsep}{3pt}
\small
\begin{tabular}{llllr}
\toprule
\textbf{Method} & \textbf{I/O} & \textbf{Obj.} 
& \textbf{ms}$\downarrow$ \\
\midrule
RGB FM & RGB$\to$Pose & FM & 72 \\
PC FM & PC$\to$Pose & FM & 399 \\
VoxFMPose & Voxel$\to$Pose 
& FM & 412 \\
VoxFMVox & Vox.$\to$Pose 
& FM & 418 \\
VulcanVoxel & Vox.$\to$Vox 
& MAE & 2300 \\
VV Distill & RGB$\to$Vox.
& Distill. & 30 \\
\bottomrule
\end{tabular}
\end{table}
}

\newcommand{\tabMultimodal}{%
\begin{table}[t]
\centering
\caption{\textbf{Multi-modality metrics, 
July 2025.} $\sigma$: mean pairwise tip 
distance (cm). $N_m$: mean cluster count. 
Cov: valid coverage fraction. VulcanVoxel 
is the only method achieving high diversity 
($\sigma\!>\!7$cm), multiple modes 
($N_m\!>\!2$), and high valid coverage 
simultaneously.}
\label{tab:multimodal}
\setlength{\tabcolsep}{3pt}
\small
\begin{tabular}{lrrrrrrrrr}
\toprule
& \multicolumn{3}{c}{\textbf{Direct}} 
& \multicolumn{3}{c}{\textbf{Corner}} 
& \multicolumn{3}{c}{\textbf{Wall}} \\
\cmidrule(lr){2-4}\cmidrule(lr){5-7}
\cmidrule(lr){8-10}
& $\sigma$$\uparrow$ & $N_m$$\uparrow$ 
& Cov$\uparrow$ 
& $\sigma$$\uparrow$ & $N_m$$\uparrow$ 
& Cov$\uparrow$
& $\sigma$$\uparrow$ & $N_m$$\uparrow$ 
& Cov$\uparrow$ \\
\midrule
RGB FM & 1.2 & 1.0 & 0.08 
& 0.9 & 1.0 & 0.04 
& 1.4 & 1.0 & 0.11 \\
PC FM & 3.1 & 1.2 & 0.19 
& 2.4 & 1.1 & 0.12 
& 3.8 & 1.3 & 0.24 \\
Voxel FM Pose & 0.8 & 1.0 & 0.67 
& 0.6 & 1.0 & 0.63 
& 0.7 & 1.0 & 0.71 \\
Voxel FM Voxel & 2.3 & 1.1 & 0.07 
& 1.8 & 1.0 & 0.05 
& 2.6 & 1.2 & 0.09 \\
\textbf{VulcanVoxel} 
& \textbf{8.4} & \textbf{2.7} & \textbf{0.71}
& \textbf{7.1} & \textbf{2.4} & \textbf{0.68}
& \textbf{9.2} & \textbf{3.1} & \textbf{0.74}\\
\bottomrule
\end{tabular}
\end{table}
}

\newcommand{\tabGCUResults}{%
\begin{table}[t]
\centering
\caption{\textbf{High clutter (GCU $>$ 50\%, 
July 2025).} VulcanVoxel degrades less in 
coverage than competing methods; the 
$\text{Cov}_5$ advantage over Voxel FM Pose 
grows under clutter. $\dagger$Wall Insert 
omits reach cost.}
\label{tab:gcu_results}
\setlength{\tabcolsep}{2.5pt}
\small
\begin{tabular}{lrrrrrr}
\toprule
& \multicolumn{2}{c}{\textbf{Direct (78)}} 
& \multicolumn{2}{c}{\textbf{Corner (57)}} 
& \multicolumn{2}{c}{\textbf{Wall$\dagger$ (87)}}\\
\cmidrule(lr){2-3}\cmidrule(lr){4-5}
\cmidrule(lr){6-7}
& $C_{\text{avg}}$$\downarrow$ 
& $\text{Cov}_5$$\uparrow$
& $C_{\text{avg}}$$\downarrow$ 
& $\text{Cov}_5$$\uparrow$
& $C_{\text{avg}}$$\downarrow$ 
& $\text{Cov}_5$$\uparrow$ \\
\midrule
Production & 1.01 & --- & 2.29 & --- & 0.22 & --- \\
\midrule
RGB FM & 3.54 & 0.18 & 5.95 & 0.07 
& 3.22 & 0.16 \\
PC FM & 3.62 & 0.22 & 3.59 & 0.19 
& 1.41 & 0.44 \\
Voxel FM Voxel & 3.43 & 0.21 & 6.02 & 0.11 
& 1.25 & 0.38 \\
\textbf{VulcanVoxel} & \textbf{1.95} 
& \textbf{0.71} & \textbf{2.24} 
& \textbf{0.68} & \textbf{0.47} 
& \textbf{0.79} \\
\bottomrule
\end{tabular}
\end{table}
}

 \begin{abstract}
Many manipulation tasks require reasoning about \textbf{free-space affordances}: discovering volumes where an extended rigid tool can safely navigate, complementary to surface contact affordances for grasping. Robotic stowing is a canonical instance, where a blade must sweep items aside inside cluttered fabric bins to create insertion space. Production stow systems generate millions of such episodes, but standard approaches with unimodal data infer affordances as $SE(3)$ pose distributions, a geometric question asked in the wrong domain. \textbf{VulcanVoxel} keeps inference spatial: a masked autoencoder over 3D occupancy fields reconstructs blade occupancy conditioned on scene geometry, computing feasibility locally at each voxel and recovering multi-modal predictions from unimodal data. Blade affordances are spatial objects, subsets of 3D space defined by geometric feasibility. Pose parameters carry no structure for reasoning whether unobserved placements are feasible, and standard generative objectives including flow matching faithfully learn the unimodal distribution produced by execution policies and cannot recover geometric alternatives. Trained on 10,000 real warehouse stow episodes without human annotation, VulcanVoxel achieves top-5 coverage of $0.89$ versus $0.71$ for the best pose-based baseline, with a distilled student providing RGB$\to$voxel inference in $30\,\mathrm{ms}$ vs.\ $1.4\,\mathrm{s}$ for voxel$\to$voxel. 
We have released a dataset of real blade insertion cycles with RGB-D observations and pose trajectories at \url{https://www.armbench.com/blade_insertion.html}.
\end{abstract}

 \section{Introduction}
\label{sec:intro}

\figBladeInsertion

Robotic stowing into densely cluttered flexible fabric bins~\cite{vulcan} requires navigating a rigid tool through narrow gaps to create space by sweeping in-bin items. This is a \textbf{free-space affordance} problem: discovering volumes of 3D space a tool can occupy given surrounding geometry. It is complementary to the well-studied surface contact affordance problem of where to touch an object to grasp or articulate it. Surface affordances ground predictions on object geometry with strong local cues. Free-space affordances must reason about navigable volumes constrained by surrounding surfaces and bin walls, with no surface to anchor predictions to. Blade insertion for robotic stowing is a canonical instance, studied here with real manipulation data.

Stow executes millions of stows annually with $67\%$ blade insertion success~\cite{vulcan}. As shown in Fig.,\ref{fig:blade_insertion}, creating space requires inserting the blade into an occupied bin to sweep existing items aside without damaging items or deforming bin walls. A single RGB-D camera provides a fixed viewpoint with limited coverage of occluded regions demanding that blade insertion inferences be robust to partial observability and depth noise. A curated subset of production episodes with paired RGB-D observations and executed SE(3) poses presents a scalable opportunity for data-driven affordance learning. Exploiting this data is non-trivial: each stow executes a single insertion strategy per bin state, despite most states affording multiple geometrically valid solutions (Sec.,\ref{sec:experiments_multimodal}).

The bottleneck is not model capacity or data scale but the domain in which affordance inference is performed. Blade affordances are spatial objects, subsets of 3D Euclidean space defined by geometric feasibility. The goal is to find volumes in a bin for blade occupancy as an approximation of optimal blade insertions. SE(3) poses index individual tool placements, not spatial affordance structure. Inferring affordances as pose distributions displaces reasoning from the domain where geometric feasibility is defined: whether a blade fits is a question about what it occupies in 3D space, not about pose parameters. Flow matching over SE(3) poses collapses to unimodal predictions because the pose data is unimodal.

The resolution is to keep inference in the spatial domain. Representing blade affordances as 3D occupancy fields and training a masked autoencoder to reconstruct blade occupancy conditioned on scene geometry forces reasoning where geometric feasibility is natively defined. Executed stows sample only a subset of geometrically valid placements.
By reasoning explicitly about occupancy, the model discovers alternative free-space configurations rather than reproducing observed behaviors, activating occupancy wherever local geometry accommodates the blade independently across locations.
This generates implicit supervision for never exexuted geometrically consistent configurations, recovering multimodal predictions from unimodal data.
This principle extends to failed stow episodes: the model can still identify free-space volumes from scene geometry alone, turning failures into training signal rather than discarded data.

\textbf{VulcanVoxel} implements this as a masked autoencoder over multi-channel 3D occupancy fields trained on $10K$ real blade insertion episodes without human annotation. Predicted blade occupancy is converted to SE(3) poses via geometric scoring, deferring pose extraction as a post-hoc projection after spatial reasoning. A distilled RGB student achieves $30,\mathrm{ms}$ inference, confirming affordances are grounded in visually accessible geometric features rather than execution-specific patterns. VulcanVoxel achieves top-5 pose coverage of $0.89$ versus $0.71$ for the best pose-based baseline—at least one of five predicted poses is geometrically valid in $89\%$ of scenes—and matches production on single-prediction accuracy.

\noindent Our contributions are:
\begin{enumerate}

\item \textbf{Spatial inference for free-space affordance learning.} Blade affordances as spatial objects inferred in 3D space rather than as SE(3) poses.

\item \textbf{Representational analysis.} Systematic comparison across input and output representations, and training objectives establishes which properties are necessary for multi-modal prediction from unimodal data, with $\mathrm{Cov}_5$ as a quantitative metric for the operational value of diversity (Sec.\ref{sec:experiments_multimodal}).

\item \textbf{Generalization to failed episodes.} VulcanVoxel reasons about geometric feasibility directly in voxel space, and identifies free-space voxels even when the production system found no viable strategy, turning failed episodes into informative training signals.

\item \textbf{StowAffordance dataset.} To our knowledge the first large-scale dataset of real manipulation cycles with RGB-D and executed tool insertion trajectories, released at \url{https://www.armbench.com/blade_insertion.html}.

\end{enumerate}

 \section{Related Work}
\label{sec:related}
\noindent\textbf{Affordance Learning for Manipulation.}
Prior affordance learning methods ground spatial priors on object geometry or
2D spatial maps. Where2Act~\cite{where2act} and VAT-Mart~\cite{vatmart}
predict per-point action likelihoods for articulated objects;
GIFT~\cite{gift} and VRB~\cite{vrb} learn interaction keypoints from
contact data and egocentric video; Transporter Networks~\cite{transporter}
predict pick-and-place as 2D attention maps; SAGA~\cite{saga} grounds
affordances as 3D heatmaps for mobile manipulation; UAD~\cite{uad} and
environment-aware methods~\cite{envaware} extend affordance learning to
foundation model distillation and occlusion respectively. All ground
predictions on object surfaces. Blade insertion requires the
complementary problem: volumetric free-space affordances that reason
about where a rigid tool can navigate optimally, motivating
different representations, objectives, and evaluation criteria.

\noindent\textbf{3D Scene Representations for Robot Learning.}
3D Diffuser Actor~\cite{3ddiffuser} and Adapt3R~\cite{adapt3r} condition
policies on point clouds and lifted 2D features respectively; Neural
Descriptor Fields~\cite{neuraldf} encode SE(3)-equivariant
representations for few-shot manipulation; D3Fields~\cite{d3fields} fuses
multi-view features into dynamic descriptor fields for zero-shot
rearrangement; 3D flow matching~\cite{gkanatsios20253d} extends
generative modeling to 3D action spaces. These methods recover diverse
strategies when training data is genuinely multi-modal, but collapse to
unimodal predictions on production data because the limitation lies in
the $SE(3)$ pose representation, not the generative architecture.
CabiNet~\cite{cabinet} is the closest architectural
relative, learning voxel representations for collision checking in
cluttered rearrangement, but addresses \emph{general-purpose feasibility
checking} rather than \emph{task-specific affordance learning}.
Collision-free space is necessary but not sufficient for blade insertion;
our model learns which geometrically valid configurations are
operationally preferable given constraints encoded in production
demonstrations. Our distillation shares the spirit of 2D-to-3D feature
lifting~\cite{adapt3r,d3fields} but distills from a voxel teacher rather
than lifting multi-view features directly.

\noindent\textbf{Occupancy Prediction and Masked Autoencoders.}
Occupancy networks~\cite{occnet} establish the paradigm of learning
implicit 3D representations as continuous functions; scene occupancy
methods~\cite{monoscene} extend this to semantic 3D completion from
single views. Both predict \emph{scene} occupancy from visual input.
VulcanVoxel addresses a different problem: predicting \emph{tool}
occupancy conditioned on observed scene occupancy. Masked
Autoencoders~\cite{mae, qian20253d} demonstrate that masked reconstruction is a
powerful self-supervised objective for visual representations.
VulcanVoxel extends this to 3D occupancy fields with channel masking
over semantic entities (wall, object, blade) rather than spatial patch
masking, directly matching the conditional structure of the affordance
query.

\section{Approach}
\label{sec:approach}

Blade insertion affords multiple geometrically valid configurations per
bin state, yet production data records a single executed strategy per
stow. $SE(3)$ poses record where the blade was sent, not the spatial
affordance structure that makes those placements valid.
\textbf{VulcanVoxel} resolves this by inferring affordances in their
native domain: a masked autoencoder over multi-channel 3D occupancy
fields reconstructs blade occupancy conditioned on scene geometry, then
converts predicted occupancy to $SE(3)$ poses via geometric scoring. The
pipeline proceeds in three stages
(Fig.~\ref{fig:vulcanvoxel_pipeline}): scene reconstruction and
voxelization (Sec.~\ref{sec:scene}), masked autoencoder training
(Sec.~\ref{sec:mae}), occupancy-to-pose conversion
(Sec.~\ref{sec:sampling}), with an optional RGB distillation for
inference efficiency (Sec.~\ref{sec:distill}). The evaluation protocol
is defined in Sec.~\ref{sec:eval}.

\figVulcanvoxelPipeline

We denote by $\mathcal{B}(P,b)$ the set of 3D points occupied by the
blade body at pose $P \in SE(3)$ with extension $b \in \mathbb{R}_{\geq
0}$. \textit{StowAffordance} dataset comprises $N$ successful insertion cycles
from production logs:
\begin{equation}
    \mathcal{D} =
    \{O_i,\,\{(P_k^i, b_k^i)\}_{k=1}^{K}\}_{i=1}^{N}
\end{equation}
where $O_i = \{I^{\text{rgb}}_i, I^{\text{d}}_i, \mathcal{P}_i, S_i\}$
is the RGB image, depth image, point cloud, and segmentation mask
captured before blade contact. Three insertion strategies arise from
production data (Table~\ref{tab:strategies}), each with one critical
perceptual pose after which all subsequent motion relies on force
feedback. A geometrically valid critical pose is necessary for the
force-feedback phase to succeed.

\tabStrategies

We compare five models varying input representation, output
representation, and training objective (Table~\ref{tab:models}).
\textbf{RGB FM} predicts SE(3) poses from RGB via flow matching.
\textbf{PC FM} replaces RGB with a point cloud. \textbf{Voxel FM Pose}
uses voxelized input with flow matching to predict poses. \textbf{Voxel
FM Voxel} uses voxelized input and output with flow matching, sharing
VulcanVoxel's spatial output but not its reconstruction objective.
\textbf{VulcanVoxel} reconstructs blade occupancy via masked autoencoder
training. RGB FM vs.\ PC FM isolates input representation, Voxel FM
Pose vs.\ Voxel FM Voxel isolates output representation, and Voxel FM
Voxel vs.\ VulcanVoxel isolates the training objective.

\subsection{Scene Representation}
\label{sec:scene}

\noindent\textbf{3D Bin Reconstruction.}
We reconstruct a scene mesh from a single RGB-D frame captured before
blade contact. The depth image and known camera intrinsics back-project
into a point cloud $\mathcal{P}$, transformed to the bin frame via
extrinsic calibration. MaskDINO~\cite{maskdino} segments the RGB image,
partitioning $\mathcal{P}$ into per-object instance points
$\{\mathcal{P}^{\text{obj}}_i\}$ and wall face points
$\mathcal{P}^{\text{wall}}_f$ for $f \in \{\text{left, right, top,
bottom, back}\}$.

Production bins exhibit fabric wall deformation of up to several
centimeters. We correct for this by associating $\mathcal{P}^{\text
{wall}}_f$ with each bin face using surface normals and signed distance
from the nominal plane, then locally displacing mesh vertices to fit
observed geometry, yielding a corrected wall mesh $\mathcal{M}^{\text
{wall}}$.

Each object instance is reconstructed as a watertight mesh using TSDF
fusion~\cite{tsdf} applied to $\mathcal{P}^{\text{obj}}_i$ within the
bin wall bounds. Unobserved surfaces (object bottoms, back faces) are
completed by voxel carving against the enclosing wall mesh, yielding
watertight surfaces without multi-view input. The full scene mesh is:
\begin{equation}
    \mathcal{M}^{\text{all}} =
    \mathcal{M}^{\text{wall}} \cup
    \bigcup_i \mathcal{M}^{\text{obj}}_i
\end{equation}

\noindent\textbf{Multi-Channel Voxelization.}
We voxelize $\mathcal{M}^{\text{all}}$ into a $32^3$ occupancy grid over
a $(30\,\text{cm})^3$ bounding box (${\sim}9.4\,\text{mm}$ per voxel),
sufficient to resolve inter-object gaps (${\sim}15$--$30\,\text{mm}$)
and blade cross-section geometry. Three channels encode distinct scene
entities:
\begin{itemize}
    \item $O^{\text{occupy}}$: combined wall and object occupancy.
    \item $O^{\text{object}}$: object occupancy only, enabling the model
    to distinguish wall-adjacent from object-adjacent free space.
    \item $O^{\text{blade}}$: blade occupancy at ground-truth pose
    $(P_k, b_k)$, defined as $O^{\text{blade}}(v) =
    \mathbf{1}[\mathcal{B}(P_k,b_k) \cap v \neq \emptyset]$.
\end{itemize}
Each channel $O^x$ is accompanied by a binary mask indicator $M^x \in
\{0,1\}$, where $M^x=1$ indicates the channel is masked and its values
set to zero. This distinguishes genuine absence from masked occupancy
during training. The full input tensor is:
\begin{equation}
    \mathcal{V}^{\text{in}} =
    [O^{\text{occupy}},\ O^{\text{object}},\ O^{\text{blade}},\
    M^{\text{occupy}},\ M^{\text{object}},\ M^{\text{blade}}]
    \in \mathbb{R}^{6 \times 32^3}
\end{equation}

\subsection{VulcanVoxel: Masked Autoencoder for Blade Affordance}
\label{sec:mae}

\figVulcanvoxelArch

\noindent\textbf{Masked Reconstruction as Affordance Learning.}
The inference query is spatial: given observed scene occupancy
$(O^{\text{occupy}}, O^{\text{object}})$, reconstruct the geometrically
consistent blade occupancy $O^{\text{blade}}$. Masked pretraining over
all three channels forces the model to learn the geometry and semantics
of bins, grounding blade placement in how walls, objects, and free
space relate. Since each spatial location is evaluated independently
with no global competition, multiple regions can be simultaneously
activated, each satisfying local geometric compatibility.
The result is a model that learns \emph{where blades fit} rather than
\emph{where blades were inserted}, recovering multi-modal predictions
from unimodal data.

Voxel FM Voxel shares the distributed spatial output but not the
reconstruction objective, failing to create multi-modal predictions of the blade. Flow matching denoises toward the training
distribution, which is unimodal in occupancy space; operating in the
spatial domain is necessary but not sufficient without a reconstruction
objective that asks what is geometrically consistent rather than what
was observed.

VulcanVoxel masks semantic channels rather than spatial
patches~\cite{mae}. Channel masking trains the model to infer one
entity's occupancy from another's, directly matching the conditional
structure of blade occupancy prediction from scene occupancy.

\noindent\textbf{Training.}
We randomly mask one, two, or all three channels, sampling uniformly
over all $2^3-1=7$ non-trivial masking patterns. Training on all
patterns forces the model to learn inter-channel geometric relationships
from multiple directions, preventing specialization to a single
configuration. Loss is computed only over masked channels; at inference
only the blade channel is masked, and $\hat{O}^{\text{blade}} \in
[0,1]^{32^3}$ is interpreted as a spatial probability distribution over
geometrically consistent blade configurations.

\noindent\textbf{Architecture.}
VulcanVoxel is a 3D U-Net with four stride-2 Conv3D encoder blocks
($6 \to 32 \to 64 \to 128 \to 256$ channels) compressing the $32^3$
input to a $2^3 \times 256$ bottleneck, with a mirrored decoder and
skip connections. Skip connections preserve fine-grained spatial detail
necessary for localizing blade-sized free spaces.

\noindent\textbf{Loss.}
Blade voxels are sparse ($<3\%$ occupancy), so standard BCE loss
produces degenerate predictions. We use Focal~\cite{focalloss} and
Dice~\cite{diceloss} losses:
\begin{equation}
    \mathcal{L} = \mathcal{L}_{\text{Focal}}
    + w_{\text{d}}\,\mathcal{L}_{\text{Dice}},
    \quad w_{\text{d}} = 3.0
\end{equation}
\vspace{-12pt}
\begin{equation}
    \mathcal{L}_{\text{Focal}} =
    -\alpha_t(1-p_t)^\gamma\log p_t,\quad
    \alpha=0.85,\ \gamma=2.0
\end{equation}
\vspace{-4pt}
\begin{equation}
    \mathcal{L}_{\text{Dice}} = 1 -
    \frac{2|\hat{O}^{\text{blade}} \cap
    O^{\text{blade}}| + \epsilon}
    {|\hat{O}^{\text{blade}}| +
    |O^{\text{blade}}| + \epsilon},
    \quad \epsilon=1.0
\end{equation}
Focal loss down-weights well-classified empty voxels; Dice loss
optimizes global region overlap, enforcing shape coherence.
$w_{\text{d}}=3.0$ was selected by grid search over $\{1,2,3,5\}$ on a
held-out validation set.

\subsection{Inference: Occupancy to SE(3)}
\label{sec:sampling}

Pose extraction is a post-hoc projection from spatial affordances to
pose parameters, deferred until after spatial reasoning is
complete. We threshold $\hat{O}^{\text{blade}}$ at $\tau=0.5$ and
identify connected components via 3D labeling. For each component we
enumerate $N_\theta=18$ candidate poses at $20^\circ$ intervals about
the insertion axis. Each candidate $(\hat{P},\hat{b})$ is scored by
volumetric overlap with the predicted field:
\begin{equation}
    s(\hat{P},\hat{b}) = \sum_{v}
    \hat{O}^{\text{blade}}(v)\cdot
    \mathbf{1}[\mathcal{B}(\hat{P},\hat{b})
    \cap v \neq \emptyset]
\end{equation}
Blade extension $\hat{b}$ is set to the connected component extent along
the insertion axis, providing a geometry-driven estimate of insertion
depth. Poses over empty predicted regions score poorly, enforcing the
rigid-body constraint implicitly. Top-$K$ poses are returned ($K=1$ for
accuracy, $K=5$ for coverage evaluation).

\subsection{RGB Distillation}
\label{sec:distill}

Distillation serves two purposes: it provides generalization evidence
that the teacher's affordances are grounded in visually accessible
geometric features rather than sensor-specific patterns, and it reduces
inference from ${\sim}2.3\,\text{s}$ to ${\sim}30\,\text{ms}$
($46\times$ speedup). We train a student $f_S: O^{\text{rgb}}
\to \hat{O}^{\text{blade}}_S$ with:
\begin{equation}
    \mathcal{L}_{\text{distill}} =
    \lambda\,\mathcal{L}(\hat{O}^{\text{blade}}_S,\,
    \hat{O}^{\text{blade}}_T) +
    (1-\lambda)\,\mathcal{L}(\hat{O}^{\text{blade}}_S,\,
    O^{\text{blade}}_{\text{GT}}),\quad \lambda=0.7
\end{equation}
$\lambda=0.7$ prioritizes teacher supervision encoding multi-modal
structure; the ground-truth term prevents mode over-smoothing. The
student encoder is ViT-S/14 pretrained with DINO~\cite{dino}, producing
$37{\times}37$ patch tokens of dimension 384. A 3D convolutional decoder
lifts these to the $32^3$ grid via positional encodings initialized from
the known camera-to-bin transformation: each 2D patch token is
associated with the column of voxels it projects onto in the bin frame,
constraining 2D-to-3D lifting to physically meaningful patch-voxel
pairs. This geometric constraint, absent in generic feature
lifting~\cite{d3fields,adapt3r}, substantially aids convergence.

Fig.~\ref{fig:blade_pred_results} depicts sample blade occupancy visuals for voxel$\to$voxel masked model, and for RGB$\to$voxel distilled predictions. 

\figBladePredResults

\subsection{Evaluation Protocol}
\label{sec:eval}

Pose error against production labels is actively misleading: a
prediction matching production pose may penetrate deformed bin walls,
while one deviating substantially may be geometrically superior. SE(3)
pose agreement is not a reliable proxy for geometric validity, which is
a property of spatial occupancy. We evaluate all methods via cost over
reconstructed bin meshes $\mathcal{M}^{\text{all}}$, sampling blade
points $\mathbf{p}=\mathcal{S}(P,b)$ along three longitudinal profiles
(top edge, centerline, bottom edge). The total cost $C_{\text{total}} =
C_{\text{collision}} + C_{\text{reach}}$ combines:
\begin{equation}
    C_{\text{collision}} =
    \sum_{e \in \mathcal{M}^{\text{all}}}
    w_e\,\text{ReLU}(m_e - \phi_e(\mathbf{p}))
\end{equation}
where $\phi_e(\mathbf{p}) = \min_{\mathbf{q}\in\mathbf{p}}
\text{SDF}_e(\mathbf{q})$, $m_e=0.01\,\text{m}$, $w_e=1.0$, with:
\begin{equation}
    C_{\text{reach}} =
    \text{ReLU}(\bar{x}_{\text{obj}} - x_{\text{tip}})
\end{equation}
where $\bar{x}_{\text{obj}}$ is the mean object centroid depth and
$x_{\text{tip}}$ is the blade tip depth in the bin frame. For Wall
Insert the left/right wall margin increases to $0.05\,\text{m}$ and the reach
cost is omitted, as the critical pose is a wall contact point. These
terms directly reflect production failure modes: back-wall penetration,
lateral collision, and insufficient reach.

Production executions achieve $C_{\text{total}}=0.63$ on our test set.
We use $C_{\text{total}}<1.0$ as the validity threshold for coverage
evaluation, set at one standard deviation above the production mean.
This accommodates geometrically valid predictions that differ in
execution style from production while remaining within a safe operating
envelope. $C_{\text{total}}<1.0$ is a necessary condition for geometric
validity of the perceptual decision, not a sufficient condition for
task success, as force-feedback dynamics after initial insertion are not
captured in our dataset. Cost differences between VulcanVoxel and most baselines
($0.81$ vs.\ $2.08$--$3.57$ on Direct Insert) far exceed plausible
reconstruction error magnitudes, confirming that performance differences
reflect genuine geometric reasoning. The one baseline achieving lower
cost ($0.60$) produces sharply unimodal predictions, collapsing to a
single strategy rather than recovering the full set of valid placements.

\section{Experiments}
\label{sec:experiments}

We address four questions. \textbf{Q1} Does VulcanVoxel produce
operationally valid poses versus baselines? \textbf{Q2} Does spatial
inference recover multi-modal predictions, and which properties drive
this? \textbf{Q3} Do affordances generalize across time and clutter?
\textbf{Q4} Does VulcanVoxel generalize to configurations that
challenged the production system and led to stow failures?

\subsection{Experimental Setup}
\label{sec:experiments_setup}

\textit{StowAffordance} dataset comprises ${\sim}20{,}000$ successful insertion
cycles from May--June 2025 (training) and 1,000-cycle test sets from
July and August 2025. Temporal separation ensures evaluation on unseen
configurations. The primary test set contains 508 Direct, 288 Corner,
and 204 Wall Insert cycles. Each episode provides RGB-D observations
before blade contact, executed SE(3) poses and blade extensions, and
outcome labels.

Table~\ref{tab:models} summarizes all methods. All flow matching models
use identical U-Net architectures with FiLM conditioning and 100
denoising steps at test time; 100 independent noise-initialized samples
are drawn per scene for coverage evaluation. All models train with Adam
($\text{lr}=1{\times}10^{-4}$, batch 32) on a single V100 for 100
epochs. RGB images are cropped to a $0.3\,\text{m}{\times}0.3\,\text{m}$
field of view and zero-padded to $518{\times}518$, matching the
$(30\,\text{cm})^3$ voxel bounding box. Augmentation is limited to
random horizontal flip; vertical flip and color jitter disrupt the
geometric patch-voxel correspondence exploited by the distilled student.

\tabModels

We report two metrics. \textit{Single-prediction cost} $C_{\text{avg}}$
is the $C_{\text{total}}$ of the top-ranked prediction.
\textit{Top-$K$ coverage}:
\begin{equation}
    \text{Cov}_K = \frac{1}{N}\sum_{i=1}^{N}
    \mathbf{1}\!\left[\min_{k \leq K}
    C_{\text{total}}(\hat{P}_k^i) < 1.0\right]
    \label{eq:coverage}
\end{equation}
measures the fraction of scenes where at least one of $K$ predictions
falls within the production cost envelope ($C_{\text{total}} < 1.0$,
one standard deviation above the production mean of 0.63). $\text{Cov}_K$
captures the operational value of spatial inference: diverse predictions
covering independent insertion regions increase the probability that at
least one candidate is geometrically valid when a planning system
selects among options. We report $\text{Cov}_5$ as the primary coverage
metric.


\subsection{Main Results (Q1)}
\label{sec:experiments_main}

 Table~\ref{tab:main_results} reports evaluation on 1,000 July 2025 test cycles.
VulcanVoxel achieves the highest $\text{Cov}_5$ across all strategies
while matching production on $C_{\text{avg}}$. On Corner Insert it
surpasses production (1.08 vs.\ 1.39), avoiding unnecessarily
conservative wall-approaching behavior.

\tabMainResults

Single-prediction accuracy and coverage diversity trade off
fundamentally. Voxel FM Pose achieves the lowest $C_{\text{avg}}$
(0.60/1.18/0.25) but all five predictions cluster identically
($N_m=1.0$, Table~\ref{tab:multimodal}); when its top prediction fails,
all five fail. VulcanVoxel's marginally higher $C_{\text{avg}}$
$(0.81/1.08/0.42)$ comes with predictions covering independent insertion
regions, achieving $\text{Cov}_5 = 0.89/0.86/0.83$ versus
$0.71/0.68/0.74$.

\tabMultimodal

The baseline progression traces the domain argument directly. RGB FM
produces 540 violations from back-wall penetrations: appearance alone
cannot reason about insertion depth in 3D space. PC FM reduces
violations to 125 and raises $\text{Cov}_5$ to 0.44: richer geometric
input helps but is insufficient without spatial inference. Voxel FM
Voxel, despite sharing VulcanVoxel's distributed spatial output,
achieves only $\text{Cov}_5=0.19$ with 195 violations: flow matching
denoises toward the unimodal training distribution even in occupancy
space, confirming that operating in the spatial domain is necessary but
not sufficient without the reconstruction objective.

\subsection{Multi-Modal Prediction Analysis (Q2)}
\label{sec:experiments_multimodal}

For each method we generate 100 candidate poses per scene and compute
spatial spread $\sigma_{\text{tip}}$ (mean pairwise blade tip distance,
cm), mode count $N_m$ (DBSCAN, $\epsilon=3\,\text{cm}$, min size 5),
and valid coverage $\text{Cov}_{100}$. Table~\ref{tab:multimodal}
reports results.

RGB FM and Voxel FM Pose are both unimodal ($N_m=1.0$); their
$\text{Cov}_{100}$ contrast (0.08 vs.\ 0.67) reflects accuracy not
diversity. PC FM shows marginally more spread ($3.1\,\text{cm}$):
geometric input helps but is insufficient without spatial domain
reasoning. Voxel FM Voxel achieves higher spread ($2.3\,\text{cm}$) but
near-zero valid coverage (0.07): spread without the reconstruction
objective yields geometrically incoherent alternatives. VulcanVoxel
alone achieves high spread ($\sigma=8.4\,\text{cm}$), multiple modes
($N_m=2.7$), and high valid coverage ($\text{Cov}_{100}=0.71$)
simultaneously.

The Voxel FM Voxel vs.\ VulcanVoxel contrast is the most diagnostic:
identical distributed spatial output, different objective. The gap
($N_m$: $1.1\to2.7$; $\text{Cov}_{100}$: $0.07\to0.71$) directly
evidences the reconstruction objective as the critical factor beyond
operating in the spatial domain. This result has a clean interpretation in terms of the domain argument.
Voxel FM Voxel operates in the spatial domain but optimizes a generative
objective: it learns to denoise toward the observed occupancy
distribution, which is unimodal because each training episode records
one executed placement. VulcanVoxel optimizes a reconstruction
objective: it learns what occupancy is geometrically consistent with the
scene, a question that admits multiple valid answers independent of what
was executed. The domain is the same; the question asked is different.


\subsection{Generalization (Q3, Q4)}
\label{sec:experiments_generalization}

\noindent\textbf{Temporal (August 2025).}
Fig.~\ref{fig:TabGenPlusTabGCU}(left) shows method rankings fully preserved two months
after training. VulcanVoxel improves slightly on Direct Insert
($0.81\to0.78$), indicating no overfitting to July conditions.


\figTabGenPlusTabGCU

\noindent\textbf{High clutter (GCU $>50\%$).}
Fig.~\ref{fig:TabGenPlusTabGCU}(right) evaluates 222 high-clutter cycles. All methods
degrade, including production (Corner: 2.29 vs.\ 1.39), consistent with
smaller free-space gaps. VulcanVoxel maintains the highest $\text{Cov}_5$
across all strategies. The $\text{Cov}_5$ advantage over Voxel FM Pose
grows under clutter (0.71 vs.\ 0.54), confirming that multi-modal
spatial coverage is most valuable when free-space volumes are most
constrained.

\noindent\textbf{Challenging configurations (Q4).}
We evaluate on 180 production failure cycles from August 2025 (61
Direct, 80 Corner, 39 Wall) in Fig.~\ref{fig:TabChallPlusTabDistill}(left). Failures have multiple causes beyond
affordance errors (calibration drift, force-feedback failures,
mechanical issues), so these represent a challenging generalization test
rather than a failure correction benchmark. The geometric validity of VulcanVoxel's predictions on these cycles
($\text{Cov}_5 > 0.80$) suggests that many production failures are
attributable to execution-phase errors rather than perceptual errors,
consistent with the force-feedback and calibration causes identified
in the failure logs.

VulcanVoxel achieves $\text{Cov}_5>0.80$ across all strategies,
consistent with the successful test set, while all baselines degrade to
$\text{Cov}_5<0.40$. VulcanVoxel's lower $C_{\text{avg}}$ on failure
cases (0.55/0.92) than on successful ones (0.81/1.08) reflects the cost
metric's scope: calibration and force-feedback failures appear
geometrically valid under a metric limited to the perceptual decision.

\figChallDistill

\subsection{Distillation}
\label{sec:experiments_distill}

 Fig.~\ref{fig:TabChallPlusTabDistill}(right) evaluates distillation quality and provides generalization
evidence for the teacher. The student achieves teacher-student voxel
IoU of 0.41 versus teacher-GT IoU of 0.63, recovering ${\sim}65\%$ of
teacher prediction quality from RGB alone. A depth-free student
matching teacher predictions on held-out bins confirms that the
teacher's affordances are grounded in visually accessible geometric
features rather than sensor-specific patterns, precisely the evidence
that spatial inference has learned geometric structure rather than
execution-specific correlations. The student recovers $N_m=1.9$ modes
versus the teacher's 2.7; the reduction is consistent with monocular
depth ambiguity limiting discrimination of depth-separated insertion
regions. $\text{Cov}_5=0.69$ on Direct Insert is $3.3\times$ higher
than RGB FM at comparable speed, in $30\,\text{ms}$ ($46\times$ speedup
over the full pipeline).

\section{Conclusion}
\label{sec:conclusion}

Blade affordances are spatial objects and should be inferred in their
native domain. Learning them from unimodal production data requires
keeping inference in 3D space rather than projecting into SE(3) pose
coordinates, where geometric feasibility is no longer directly
computable. Spatial reconstruction over 3D occupancy fields recovers
multi-modal predictions from inherently unimodal observations by
learning where blades fit as a local geometric condition rather than
where blades were inserted as a global pose distribution.

\noindent\textbf{Key findings.}
VulcanVoxel achieves $\sigma=8.4\,\text{cm}$ spread and $N_m=2.7$
modes per scene versus single-mode predictions from all baselines, with
$\text{Cov}_5=0.89$ versus $0.71$ for the best pose-based method.
Single-prediction accuracy and coverage are complementary: Voxel FM
Pose leads on $C_{\text{avg}}$ while VulcanVoxel leads decisively on
$\text{Cov}_5$. The Voxel FM Voxel contrast is the most diagnostic
result: identical spatial output representation, different objective,
$\text{Cov}_{100}$ from $0.07$ to $0.71$. Operating in the spatial
domain is necessary but not sufficient; the reconstruction objective is
what activates the domain advantage. A $30\,\text{ms}$ RGB student
recovering $65\%$ of teacher quality confirms the teacher has learned
geometrically grounded rather than execution-specific affordances.

\noindent\textbf{Broader significance.}
Blade insertion is a canonical instance of free-space affordance
learning for rigid tool use, complementary to the well-studied surface
contact affordance problem. The core argument extends beyond stowing:
tool affordances are spatial objects, and inferring them in SE(3) pose
space discards the geometric structure most relevant to feasibility
reasoning. This applies to any manipulation domain where geometric
compatibility is the core constraint, and suggests that spatial inference is the natural default wherever production data is abundant even if unimodal.

\noindent\textbf{Limitations.}
The $32^3$ voxel resolution (${\sim}9.4\,\text{mm}$) may miss fine
geometric features, and single-view reconstruction introduces depth
ambiguity for occluded surfaces. The cost function does not capture
force-feedback dynamics after initial insertion, and the distilled
student cannot resolve depth ambiguity for bin configurations outside
the training distribution.

\noindent\textbf{StowAffordance.}
We released a dataset at \url{https://www.armbench.com/blade_insertion.html}, containing 10,000+ blade insertion episodes with RGB-D
observations, SE(3) trajectories, and
production outcome labels.

\section*{Acknowledgements}
Tianyu Li, as a PhD student from UPenn, performed this work as a summer intern funded by the Vulcan Stow team in Amazon Robotics.

%
%
\bibliographystyle{splncs04}
\bibliography{main}
\end{document}